\documentclass{article}

 \usepackage[dblblindworkshop, final]{neurips_2025}

\usepackage[utf8]{inputenc} 
\usepackage[T1]{fontenc}    
\usepackage{hyperref}       
\usepackage{url}            
\usepackage{booktabs}       
\usepackage{amsfonts}       
\usepackage{nicefrac}       
\usepackage{microtype}      
\usepackage{xcolor}         

\usepackage{amsmath}
\usepackage{graphicx}
\usepackage{caption}
\usepackage{cleveref}
\usepackage{float}
\usepackage{multirow} 
\hypersetup{
    colorlinks,
    linkcolor={red!50!black},
    citecolor={blue!50!black},
    urlcolor={blue!80!black}
}

\title{Interpreting ResNet-based CLIP \\ via Neuron-Attention Decomposition}
\workshoptitle{Mechanistic Interpretability}

\author{%
  Edmund Bu \\
  UC San Diego\\
  \texttt{ebu@ucsd.edu} \\
  \And
  Yossi Gandelsman \\
  UC Berkeley \\
  \texttt{yossi\_gandelsman@berkeley.edu} \\
}

\begin{document}

\maketitle

\begin{abstract}
    We present a novel technique for interpreting the neurons in CLIP-ResNet by decomposing their contributions to the output into individual computation paths. More specifically, we analyze all pairwise combinations of neurons and the following attention heads of CLIP's attention-pooling layer. We find that these neuron-head pairs can be approximated by a single direction in CLIP-ResNet's image-text embedding space. Leveraging this insight, we interpret each neuron-head pair by associating it with text. Additionally, we find that only a sparse set of the neuron-head pairs have a significant contribution to the output value, and that some neuron-head pairs, while polysemantic, represent sub-concepts of their corresponding neurons. We use these observations for two applications. First, we employ the pairs for training-free semantic segmentation, outperforming previous methods for CLIP-ResNet. Second, we utilize the contributions of neuron-head pairs to monitor dataset distribution shifts. Our results demonstrate that examining individual computation paths in neural networks uncovers interpretable units, and that such units can be utilized for downstream tasks.
\end{abstract}

\section{Introduction}

Interpreting the hidden components in pre-trained deep neural networks, by tracing their contribution to the model output, allows us to detect model limitations and useful sub-computations that can be repurposed for multiple downstream tasks (\cite{LLMbio}, \cite{sharkey2025openproblemsmechanisticinterpretability}). Recently, such approaches were applied to CLIP - a widely used class of image encoders (\cite{textspan}, \cite{splice}, \cite{SecondOrderLens}). These approaches unlocked various capabilities - discovery of spurious correlations in the model output, automated generation of adversarial attacks, and even reuse of the interpreted model components for segmentation.

While CLIP interpretability work showed promising results for vision transformer-based variants (CLIP-ViT), such existing methods do not readily extend to the ResNet counterparts (CLIP-ResNet). Most of these existing methods rely on a decomposition of the output into a sum of per-layer contributions, which is not possible for CLIP-ResNet as, despite its additive residual connections, each layer is followed by a non-linearity. Moreover, CLIP-ViT interpretability methods rely on the model attention blocks and a special class token, while CLIP-ResNet models have convolutions and final attention pooling instead. These architectural differences make existing methods not applicable.

To address this gap, we introduce a new approach that provides a fine-grained decomposition of CLIP-ResNet model outputs into individual contributions of neurons in the last layers and the following attention-pooling heads. More specifically, we show that the output of CLIP-ResNet is a sum over computation paths, ranging from individual neurons in the last layer, in parallel through all the attention heads in the attention pooling layer, to the output. As each such neuron-head contribution lives in the joint image-text space, it can be compared and interpreted via text.

We analyze the contributions of each neuron-head pair to the output and find that, unlike CLIP-ResNet's individual neurons, neuron-head pairs can be approximated by a \textit{single direction} in the joint image-text embedding space. Additionally, we find that a fixed sparse set of neuron-head pairs comprises most of the output value (mean-ablating a fixed bottom 80\% of neuron-head pairs decreases ImageNet classification accuracy by only 5\%). We run analogous experiments for neuron contributions and find these properties to be unique to neuron-head pairs. Furthermore, we discover that some neuron-head pairs encompass sub-concepts of the concepts that the corresponding neuron represents (e.g., a `butterfly' neuron can be decomposed into a `butterfly clothing' neuron-head pair and other sub-concepts).

We leverage our findings for two applications: semantic segmentation and monitoring dataset distribution shifts. Given the approximation of each neuron-head pair by a single direction, we can rank each pair by its similarity to text representations. We apply this to associate each class with a set of $k$ neuron-head pairs, and use this association for dense semantic segmentation. Notably, we achieve a 15\% relative improvement in mIoU over previous methods for training-free semantic segmentation using CLIP-ResNet. We also conduct a case study on monitoring dataset distribution shift, in which we show that neuron-head pair contributions closely track the ground truth of the concepts they represent, and are different between the classes.

In summary, we propose neuron-attention decomposition as an improved interpretability method to automatically label the components of CLIP-ResNet's vision encoder with text. We find evidence that our decomposition is favorable in comparison to decomposing solely neurons or solely attention heads, and apply this decomposition to two relevant applications. This demonstrates the viability of examining fine-grained computation paths for studying and enhancing model capabilities.

\section{Related Work} 
\label{related-work}

\textbf{Contrastive vision-language models.} 
Models like CLIP (\cite{radford2021learningtransferablevisualmodels}) and its variants (\cite{Jia2021ScalingUV}, \cite{zhai2023sigmoidlosslanguageimage}) are trained on massive web-based datasets of images and their captions to learn meaningful image representations. This pre-training enables zero-shot capabilities for downstream tasks like OCR, geolocalization, and classification (\cite{wortsman2023openclip}). These models are also used as backbones for other systems such as LLaVA (\cite{llava}), 3-D learning~(\cite{pointclip}), and image generation (\cite{ramesh2021zeroshottexttoimagegeneration}, \cite{rombach2022high}).

\textbf{Interpreting vision models.} 
Mechanistic interpretability is a field of research that seeks to understand the inner workings of neural networks by analyzing fundamental model components and computation paths. Early mechanistic interpretability discoveries for vision models include the attribution of high-level concepts to intermediate model neurons (\cite{network-dissection}), and curve detectors and circuits (\cite{circuits-thread:}). Similar to us, a body of work aims to automatically label vision model components with text (\cite{milan}, \cite{bills2023language}, \cite{LE}).

\textbf{Interpreting CLIP.} 
Several recent works investigate CLIP's embedding space with techniques like sparse coding (\cite{splice}) and factor rotation (\cite{quantifying_structure}), aiming to identify human-interpretable concepts. Most closely to us, previous work analyzing CLIP-ViT's output decomposition (\cite{SecondOrderLens}, \cite{textspan}) by utilizing the additive linearity of the residual stream to examine contributions of individual components to the embedding space, and relied on the joint output image-text space to interpret such components with text. Differently from these methods, we focus on ResNet-based models, for which the contributions of the early layers are not additive to the output (due to a ReLU non-linearity following each residual connection). However, we \textit{are} able to interpret CLIP-ResNet's last convolutional block and its attention pooling, where we propose a more fine-grained view of its inner workings by analyzing neuron-head pairs.

\section{Methodology}
\label{methodology}

We start by presenting CLIP-ResNet's architecture and deriving the decomposition of its image representation. We use this decomposition in later sections to interpret the contributions of individual pairs of neurons and attention heads.

\subsection{CLIP-ResNet preliminaries}
\label{preliminaries}

\textbf{Contrastive pre-training.}
CLIP is trained via a contrastive loss that aligns the representations of its image encoder $M_\text{image}$ with the representations of its text encoder $M_\text{text}$ in a shared image-text latent space $\mathbb{R}^d$. Specifically, over massive web-based datasets, the two encoders are trained together to maximize the cosine similarity for matching image-text pairs $(I, t)$:
\begin{equation}
    \mathrm{sim}(I, t) = \frac{\langle M_\text{image}(I), M_\text{text}(t) \rangle}{\|M_\text{image}(I)\|_2 \, \|M_\text{text}(t)\|_2}.
\end{equation}

\paragraph{Zero-shot classification.}
To perform image classification, each class name $c_j$ is mapped to some template (e.g., ``A photo of a \{class\}'') and encoded by the text encoder as $M_\text{text}(\mathrm{template}(c_j))$ (for simplicity, we will omit the template notation). The classification prediction for an image $I$ is the class $c_j$ whose text representation $M_\text{text}(c_j)$ is most similar to the image representation $M_\text{image}(I)$.

\textbf{CLIP-ResNet.}
The CLIP-ResNet image encoder is a traditional ResNet network~(\cite{he2016residual}), composed of sequential \textit{residual blocks} with an average pooling replaced by attention pooling (\cite{radford2021learningtransferablevisualmodels}). While CLIP is often trained with ViT as the image encoder backbone, CLIP-ResNet is competitive in performance across various benchmarks. However, the internal mechanisms of CLIP-ResNet are underexplored in comparison to CLIP-ViT, and existing methods are not applicable due to architectural differences.

\textbf{CLIP-ResNet architecture.}
CLIP-ResNet's residual stream is not linear, as a nonlinear ReLU activation follows the additive residual connection within each residual block. Thus, we focus only on the final residual block's input to the attention pooling layer, which we \textit{can} decompose linearly. Formally, given an input image $I$, let $Z(I)$ be the output of the last convolutional layer in the model, and let $Z^\prime(I)$ be $Z(I)$ after prepending a class token and adding positional embedding. CLIP-ResNet's image representation is the $Z^\prime(I)$ with an attention pooling applied to it:
\begin{equation} 
    M_\text{image}(I) = \mathrm{AttnPool}(Z^\prime(I)) .
\end{equation}
More specifically, the dimensionality of $Z(I)$ is ${C \times H' \times W'}$ where $C$ is the number of \textit{neurons} (post-ReLU per-location activations in the final convolutional block) and $H'$ and $W'$ are the spatial feature map dimensions. To form $Z^\prime(I) \in \mathbb{R}^{(K+1)\times C}$, $Z(I)$ is first flattened into $K=H'W'$ image tokens $\{z_i\}_{i \in \{1, \ldots,K\}}, z_i \in \mathbb{R}^C$, then a \textit{class token} $z_0 = \frac{1}{K}\sum_{i=1}^{K} z_i$ is prepended to this sequence, and finally, a learned positional embedding is added to all $K+1$ tokens. Attention pooling is implemented as a standard transformer multi-head attention module, with the exception being that the class token is used as the sole output $M_\text{image}(I)$.

\subsection{Decomposition into neurons, heads, and tokens}
\label{decomposition}

\textbf{Decomposition into attention heads and tokens.}
Following \cite{elhage2021mathematical} and leveraging the fact that only the class token is returned by attention pooling, we can write the image representation as a sum over $H$ attention heads of the attention-pooling layer and $K+1$ tokens:
\begin{equation}
    M_\text{image}(I) = \mathrm{AttnPool}([z_0, \ldots, z_{K}]) = \sum_{h=1}^H \sum_{i=0}^{K} a_i^{h}(I) \, z_i\,W_{VO}^h
    \label{head_token}
\end{equation}
where $W_{VO}^h \in \mathbb{R}^{C \times d}$ are transition matrices (the OV matrices) and $a_i^h$ is a weight that denotes how much the class token attends to the $i$-th token ($\sum_{i=0}^{K} a_i^h = 1$). 

\begin{figure}[t]
    \centering
    \begin{minipage}[c]{0.48\linewidth}
        \centering
        \begin{tabular}{l c}
            \toprule
            PC(s) & Accuracy (\%) \\
            \midrule
            (Baseline) & 70.7 \\
            $\{\hat{r}^{n, h}\}$ & 70.7 \\
            $\{\hat{r}^{n}_1\}$ & 66.9 \\
            $\{\hat{r}^{n}_1, \hat{r}^{n}_2\}$ & 69.0 \\
            $\{\hat{r}^{n}_1, \hat{r}^{n}_2, \hat{r}^{n}_3\}$ & 69.7 \\
            $\{\hat{r}^{n}_1, \hat{r}^{n}_2, \hat{r}^{n}_3, \hat{r}^{n}_4\}$ & 70.0 \\
            \bottomrule
        \end{tabular}
        \captionof{table}{\textbf{Accuracy for reconstruction from principal components.} Unlike neuron representations, neuron-head representations match the baseline accuracy using only one direction for reconstruction.}
        \label{table-PCs}
    \end{minipage}
    \hfill
    \begin{minipage}[c]{0.48\linewidth}
        \vspace{0pt}
        \centering
        \includegraphics[width=0.9\linewidth]{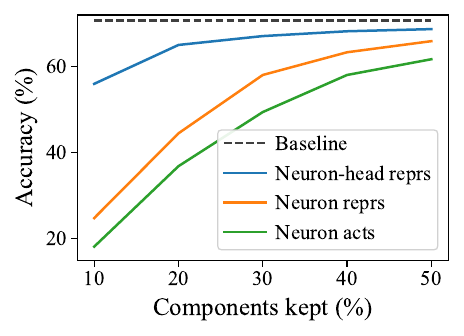}\vspace{-0.5em}
        \captionof{figure}{\textbf{Mean-ablation accuracy.}\\ Mean-ablating all but a fixed set of components shows there is a sparse set of neuron-head pairs that is very significant.}
        \label{fig-mean-ablate}
    \end{minipage}
\end{figure}

\textbf{Decomposition into neuron-head pairs.}
Each row of the $W_{VO}^h$ matrix corresponds to one neuron. That means, for any given head and token, we can rewrite: 
\begin{equation}
    z_i \, W_{VO}^h  = \sum_{n=1}^C z_i \, W_{VO}^{n, h} 
\end{equation}
where the superscript $n$ denotes the $n$-th row of $W_{VO}^h$. Substituting into ~\eqref{head_token} and swapping the sum yields:
\begin{equation}
    M_\text{image}(I) = \sum_{n=1}^C \sum_{h=1}^H \sum_{i=0}^K r^{n, h}_i(I), \quad r^{n, h}_i = a_i^{h}(I) \, z_i \, W_{VO}^{n, h}  .
    \label{r_eq}
\end{equation}
$r^{n, h}_i$ denotes the $d$-dimensional contribution to the output from the class token's attention to token $i$ through head $h$, projected by the row of $W_{VO}^h$ that corresponds to neuron $n$.

The image representation $M_\text{image}$ lives in the joint image-text embedding space. Therefore, each neuron-head contribution $r^{n, h}$ obtained by summing over tokens, and each neuron contribution $r^n$, obtained by summing over tokens \textit{and} heads, lives in the same image-text space. This allows us to compare them to text. We will use this property for automatically labeling neurons and neuron-head pairs in later sections. Summing over the neuron dimension, instead, gives the decomposition $r^{h}_i$, which corresponds to the spatial patches at locations $i \in \{0, \ldots, K\}$. This allows us to compute the similarity to text for each image location represented in the decomposed class token. We use this property to form our per-token segmentation map in~\Cref{segmentation}.

\textbf{Comparison to existing decompositions.} 
Previous work on CLIP-ViT has studied the direct effect of decomposition across attention heads and tokens~(\cite{textspan}), analogous to our decompositions $r^{h}_i$ and $r^h$, as well as the second-order effect of decomposition into neurons~(\cite{SecondOrderLens}), analogous to our decomposition $r^n$. We show that our approach overcomes two main limitations of these methods, when applied to CLIP-ResNet: First, there is a small number of attention heads relative to possible concepts that CLIP learns, meaning that each head encodes multiple concepts, ultimately making them less interpretable (\cite{incidental-poly}). Second, the second-order effect of neurons disregards the independent nature of attention heads in the multi-head attention layer, meaning $r^n$ encompasses all paths through the network -- which, as we will show next, captures multi-dimensional conceptual structure that is yet again difficult to interpret.

\vspace{-0.5em}
\section{Analysis of individual components}
\vspace{-0.5em}

We quantitatively evaluate contributions of individual components in our neuron-attention decomposition, and show their benefits over other decompositions -- these contributions are approximately rank-1 and sparse. Additionally, we qualitatively investigate the top-activating images for selected components and discover that neuron-head pairs tend to represent sub-concepts of the concepts represented by the corresponding neuron. We later use these observations to produce semantic segmentation~(\Cref{segmentation}) and monitoring distribution shift~(\Cref{Case-studies}).

\begin{figure}[t]
    \centering
    \begin{minipage}[c]{0.48\linewidth}
        \vspace{0.5em}
        \centering\includegraphics[width=0.98\linewidth]{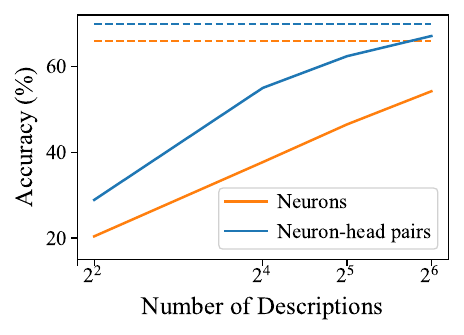}
        \vspace{-0.5em}
        \captionof{figure}
        {\textbf{Reconstruction from sparse text representations.} We compare sparse text decompositions for neurons and neuron-head pairs while varying description set sizes.}
        \label{fig-sparse-coding}
    \end{minipage}
    \hfill
    \begin{minipage}[c]{0.48\linewidth}
        \centering
        \includegraphics[width=0.98\linewidth]{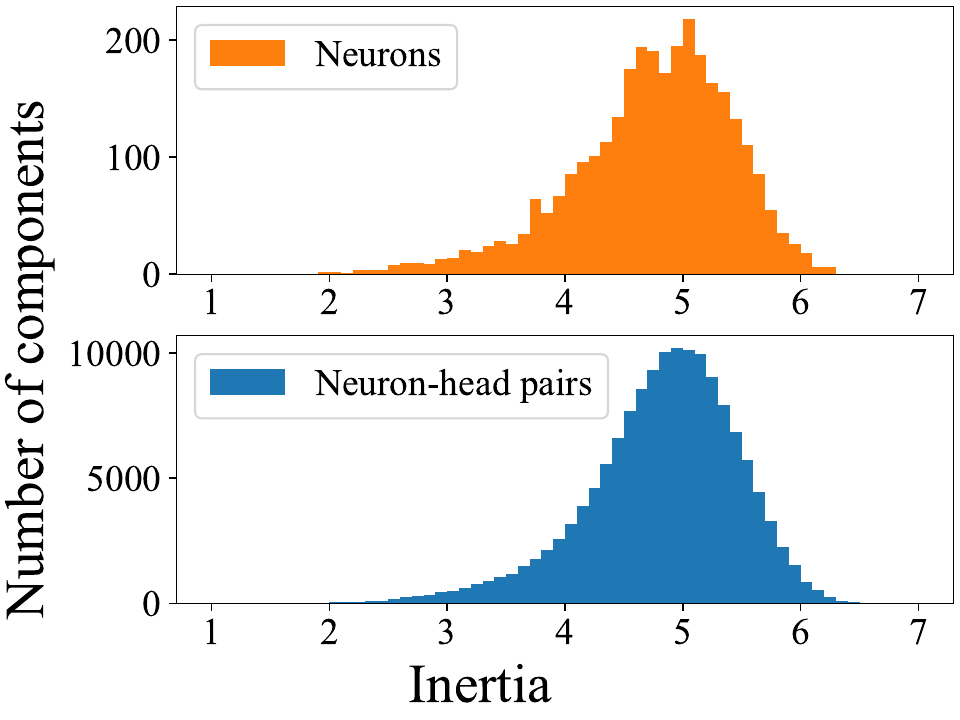}
        \captionof{figure}{\textbf{Inertia as a proxy to compare polysemanticity}. We display bins at 0.1 intervals and compute cluster metrics on each component's top ten images by contribution norm from $\mathcal{D}$.}
        \label{fig-poly}
    \end{minipage}
    
\end{figure}

\subsection{Quantitative analysis}
\label{quant-analysis}
\vspace{-0.3em}

We study the properties of $r^{n,h}$ and $r^{n}$. We show that individual $r^{n, h}(I)$ can be approximated by a single direction in the image-text space, while approximating $r^n(I)$ with only one linear direction causes a significant drop in reconstruction fidelity. Moreover, we find that neuron-head pairs are contributing more sparsely than neurons-only -- mean-ablating all but a subset of each component shows a sparser set of top-contributing neuron-head pairs in comparison to neurons.

\textbf{Experimental setting.} 
We measure the performance of zero-shot classification on the ImageNet (\cite{ImageNet}) validation dataset after various ablations to quantify the resulting change in the representation. We collect $r^{n,h}(I)$ and $r^n(I)$ contributions over the set $\mathcal{D}$, which is comprised of 1000 images from the ImageNet test dataset. We use our collected representations to compute singular value decomposition, from which we obtain our principal components, and to compute mean contributions, for our mean-ablation (\cite{nanda2023progressgrok}) experiments. Additionally, after obtaining each neuron-head pair's principal component, we further decompose it using a sparse coding technique and evaluate its sparsity-reconstruction tradeoff. We perform all experiments using OpenAI's CLIP-RN50x16, and present additional results for RN50x64 in \Cref{appendix-other-model}.

\textbf{Neuron-head contributions are one-dimensional.}
We find that $r^{n, h}(I)$ can be approximated by a single direction $\hat{r}^{n, h}$ in the joint embedding space. We approximate $r^{n, h}(I)$ for any given image $I$ with $x^{n, h}(I) \, \hat{r}^{n, h} + b^{n, h}$, where $x^{n, h}(I)$ is the coefficient obtained from the projection norm of $r^{n, h}(I)$ onto $\hat{r}^{n, h}$, and $b^{n, h}$ is the mean of all $r^{n,h}(I)$ over all images $I \in\mathcal{D}$. As shown in~\Cref{table-PCs}, replacing $r^{n, h}(I)$ with this approximation results in no decrease in reconstruction quality, as measured by downstream ImageNet classification accuracy.

\textbf{Neuron-head contributions are sparse.} 
Keeping only 20\% of neuron-head contributions, computed over ImageNet (while mean-ablating the rest), results in only a $\sim5\%$ decrease in classification accuracy. We sort the $C\times H$ neuron-head pairs by the mean of their top percentile norms over $\mathcal{D}$, and observe the same high-scoring pairs $\mathcal{P}^*$ tend to be consistently important across the dataset. Decomposing $M_\text{image}(I)$ into individual $r^{n, h}(I)$ contributions, we keep only the $r^{n, h}(I)$ whose neuron-head pairs $(n, h)$ are in $\mathcal{P}^*$ and mean-ablate the rest. We construct $\mathcal{P}^*$ with varying top percentage norms (at 10\% increments) and measure the resulting classification accuracy on ImageNet validation in~\Cref{fig-mean-ablate}. As shown, most of the output value can be recovered from a sparse set of neuron-head pairs.

\textbf{Sparse text-based decomposition of neuron-head directions.}
Following \cite{SecondOrderLens}, we use orthogonal matching pursuit (\cite{Pati1993OrthogonalMP}) to further decompose each $\hat{r}^{n, h}$ direction into sparse text components. Formally, we use a sparse set of text components $\{t_j\}_{j \in\{1, m\}}$ to approximate $\hat{r}^{n, h} \approx \sum_{j=1}^m \gamma^{n, h}_j \, M_\text{text}(t_j)$, where $\gamma^{n, h}_j$ is a non-zero scalar coefficient. In our experiments, the initial pool of text descriptions is composed of the 30,000 most common English words. We vary the sparse set size $m$ and present classification accuracy performance on ImageNet, after replacing $\hat{r}^{n, h}$ with our sparse approximation in \Cref{fig-sparse-coding}. As shown, using $m=64$ text descriptions for neuron-head pairs surpasses the neuron baseline in reconstruction accuracy, exemplifying the benefits of our fine-grained decomposition.

\textbf{Comparison to neuron-only contributions.}
We repeat the two experiments above for neuron-only decomposition. To include the variance explained by multiple principal components, we simply reconstruct from a set $\{\hat{r}^n_1, \hat{r}^n_2, \ldots\}$, where $\hat{r}^n_k$ is the $k$-th principal component. We use $\hat{r}^n_1$ for all text-based sparse decomposition experiments. Compared to our findings above for neuron-head pairs, the first principle component $\hat{r}^n_1$ explains much less variance -- it reconstructs with a $\sim4\%$ drop in accuracy from the baseline. Additionally, the contributions are less sparse -- our mean-ablation experiment shows a nearly $38\%$ accuracy difference between top $10\%$ neurons and top $10\%$ neuron-head pairs (\Cref{fig-mean-ablate}). We also show results for mean-ablation of neuron activations, which performs even worse. Finally, as shown in~\Cref{fig-sparse-coding}, neurons show less sparsity in reconstruction from text components.

\subsection{Qualitative analysis}

\begin{figure}[t]
    \centering
    \includegraphics[width=0.95\linewidth]{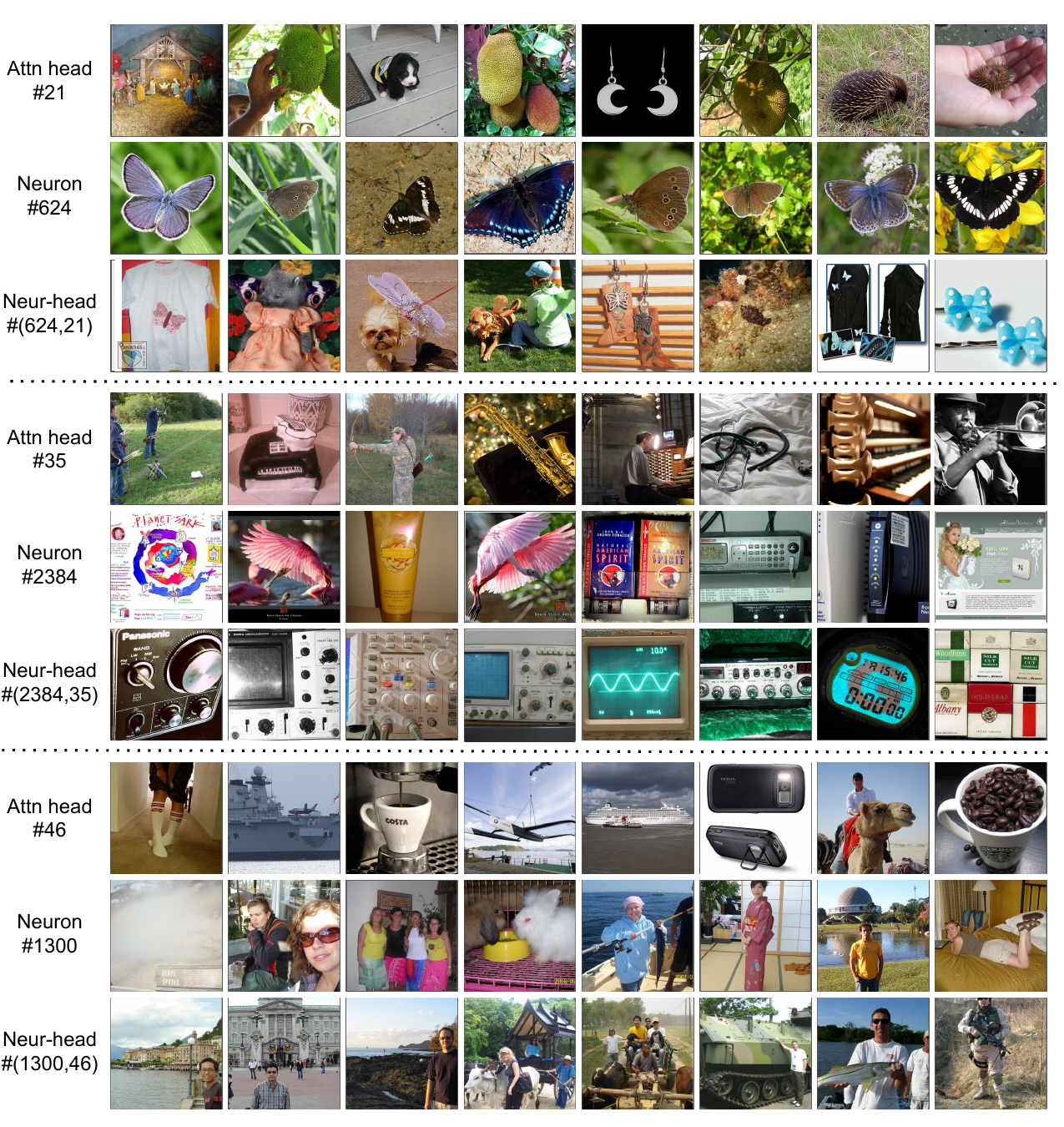}
    \vspace{-1.em}\caption{\textbf{Images with largest contribution norm for attention heads, neurons, and neuron-head pairs.} We present the top images from ImageNet validation set. Neuron-head pairs correspond to specific subcategory concepts of neurons (e.g., `butterfly \textit{clothing}' in row 3) and similar concepts to their neurons (e.g., `router' for neuron $\#2384$ and `people' for neuron $\#1300$)}
    \label{fig-image-retrieval}
\end{figure}

We qualitatively analyze the images $I$ across the ImageNet validation dataset that provide the highest contribution $r^h(I)$, $r^n(I)$, or $r^{n, h}(I)$ in norm. In~\Cref{fig-image-retrieval}, we select pair $\#(624, 21)$ as the pair with the top cosine similarity to $M_\text{text}(``butterfly'')$, and choose the other two pairs from four randomly selected pairs. We note that, as shown in the figure, the top-activating images for a given neuron-head pair appear similar to those of its corresponding neuron, but not its corresponding head. Notably, the neuron $\#624$ is most active on images that represent the `butterfly' concept, and the neuron-head pair $\#(624, 21)$ is most active for the \textit{sub-concept} `butterfly clothing'. These examples highlight the ability of neuron-attention decomposition to isolate semantically meaningful directions in CLIP's embedding space.

\begin{figure}[t]
\small
    \vspace{0mm}
    \centering
    \begin{tabular}{l l}
        \toprule
        \textbf{Components} & \textbf{Text descriptions }\\
        \midrule\midrule
        Attention head \#21 & ``slr", ``vantage", ``jetta"\\
        \midrule
        Neuron \#624 & ``wings" ($+1.73$),  ``butterfly" ($+1.69$), ``kite" ($+1.20$) \\
        \midrule
        Neuron-head \#(624, 21) & ``butterfly" ($+2.64$), ``roses" ($-1.17$), ``grizzlies" ($+0.91$) \\
        \midrule
        \midrule
        Attention head \#35 & ``musicians", ``motorcycles", ``archery" \\
        \midrule
        Neuron \#2384 & ``kobe" ($+1.00$), ``alfa" ($-0.98$),  ``redmond" ($+0.91$) \\
        \midrule
        Neuron-head \#(2384,35) & ``routers" ($+2.10$), ``tutorials" ($-0.99$),  ``spur" ($-0.94$) \\
        \midrule
        \midrule
        Attention head \#46 & ``salford", ``jcpenney",  ``chattanooga" \\
        \midrule
        Neuron \#1300 &  ``smileys" ($+1.60$),  ``masterpieces" ($+1.31$),  ``affiliated" ($+1.16$) \\
        \midrule
        Neuron-head \#(1300,46) & ``cnn" ($+1.26$),  ``cto" ($-1.25$),``varsity" ($-0.99$)  \\
        \bottomrule
    \end{tabular}
    \captionof{table}{\textbf{Sparse text-based decomposition examples for~\Cref{fig-image-retrieval} components.} We select from the descriptions detailed in~\Cref{quant-analysis} with sparsity $m=64$ and use TextSpan (\protect\cite{textspan}) to obtain descriptions for each attention head.}
    \label{table-sparse-decomposition}
\end{figure}
\begin{figure}[t]
    \centering
\vspace{-1.25em}\includegraphics[width=1.\linewidth]{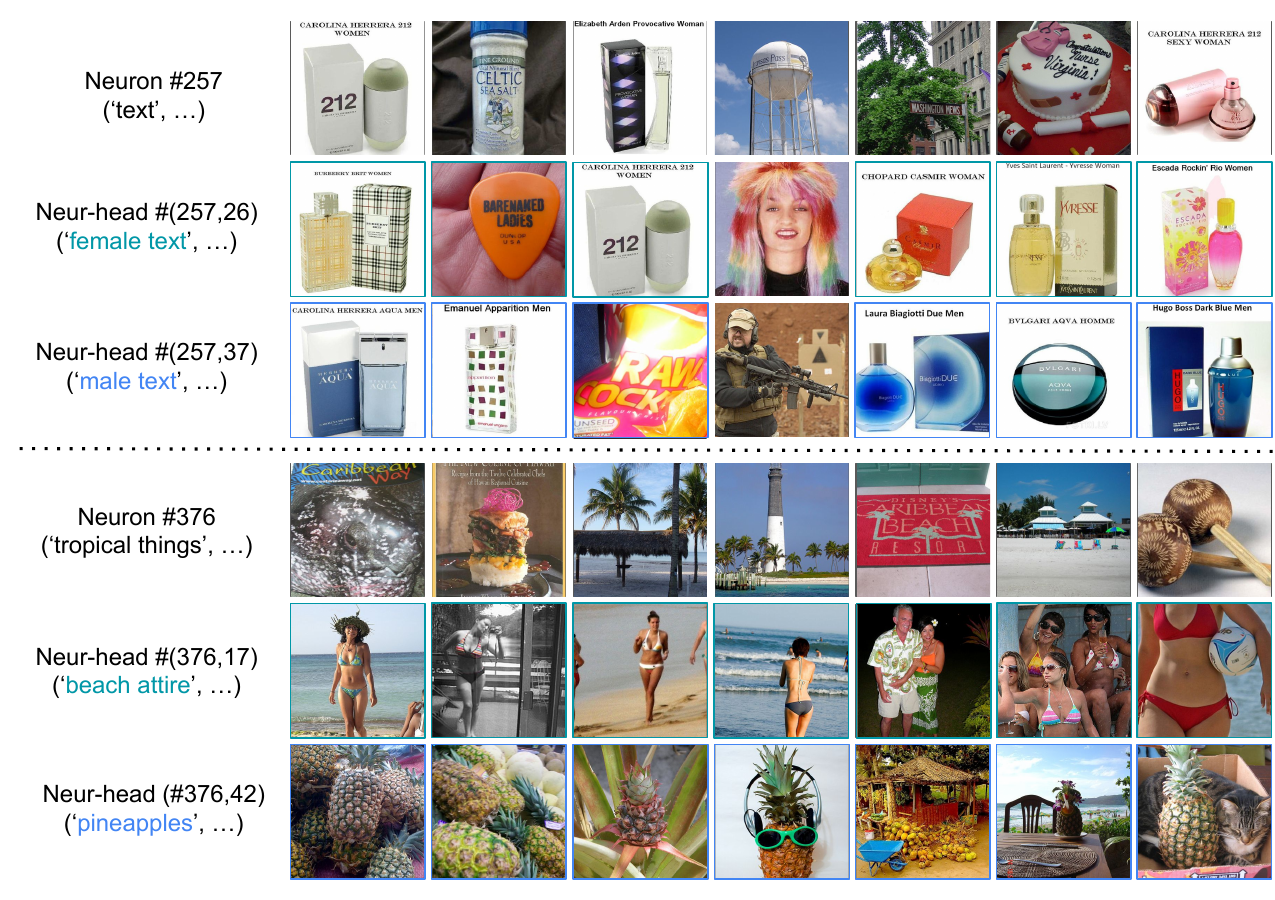}
    \vspace{-1.5em}\caption{\textbf{Sub-concept neuron-head pairs for select neurons.} We again present top images by contribution norm and depict sub-concept relationships for two different neurons.}
    \label{fig-subconcept}
\end{figure}

\textbf{Neuron-head pairs remain polysemantic.}
We observe that neuron-head pairs remain polysemantic (for instance, in~\Cref{fig-image-retrieval}, the 6th image for pair $\#(624, 21)$ and the 9th image for pair $\#(2384, 35)$ show polysemanticity). We collect the top ten images by contribution norm, from $\mathcal{D}$, for each neuron and neuron-head pair, and present the inertia of the normalized image embeddings (see \Cref{fig-poly}). Lower inertia implies a tighter cluster and less polysemanticity. By raw count, there are far more less-polysemantic neuron-head pairs than neurons, an advantage brought about by neuron-attention decomposition.

\begin{figure}[t]
    \centering
    \vspace{0pt}
    \centering
    \includegraphics[width=0.95\linewidth]{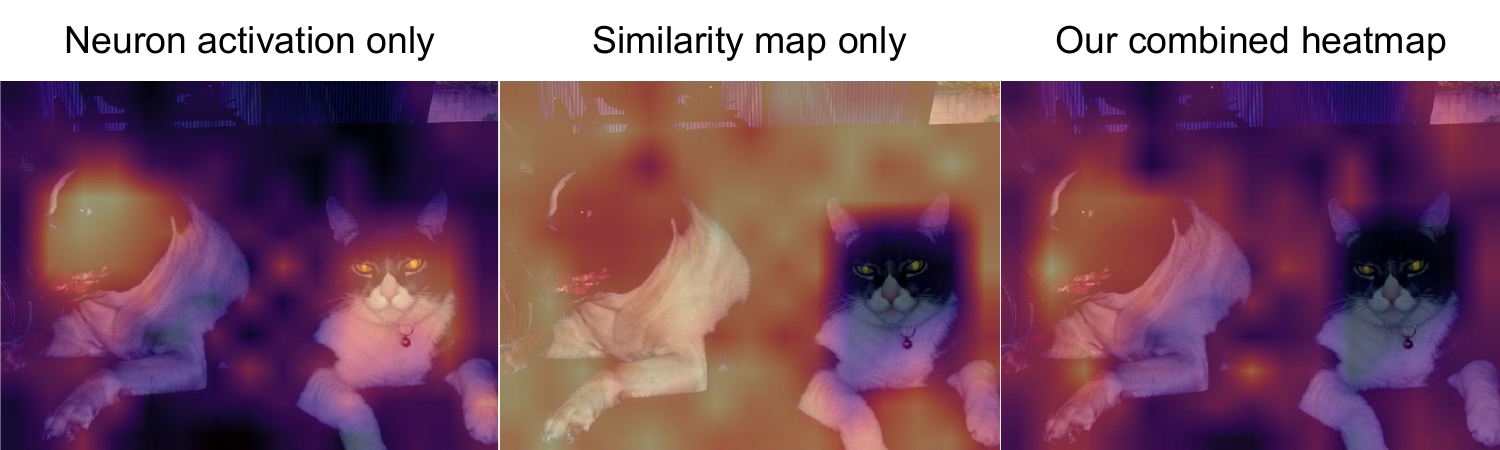}
        \captionof{figure}{\textbf{Class segmentation maps for `dog'.} Our method of multiplying two heatmaps together mitigates the failure modes of either one. Here, those failure modes are 1) the focus on the cat shown by $\sum_{r=1}^k Z^{n_r}(I)$ and 2) the negative localization shown by $\sum_{r=1}^k L_\text{sim}^{h_r}(I)$.}
    \label{fig-seg-visualization}
\end{figure}

\textbf{Neuron-head concepts are sub-concepts of neuron concepts.}
Inspired by the `butterfly clothing' pair in~\Cref{fig-image-retrieval}, we search for more image retrieval examples of sub-concept neuron-head pairs. We detail this process in~\Cref{appendix-subconcept} and present two examples in~\Cref{fig-subconcept} and the rest in the appendix. We find several sub-concept pairs in all of the randomly selected neurons, demonstrating that our neuron-attention decomposition captures fine-grained sub-concepts.

\section{Applications}
\vspace{-0.5em}
\subsection{Semantic segmentation}
\label{segmentation}

We use the observation from above to repurpose CLIP for training-free semantic segmentation. Existing interpretability-based approaches used for segmentation (\cite{textspan}, \cite{SecondOrderLens}, \cite{helbling2025conceptattention}) focus on ImageNet-Segmentation~\cite{guillaumin2014imagenet} or other single-class tasks. We aim to show the viability of interpretability for multi-class segmentation tasks, validating our approach in the paradigm of training-free CLIP semantic segmentation (\Cref{appendix-segmentation}). Unlike related works in this area, we do not alter any computation within CLIP (i.e., we do not use self-self attention). We use CLIP's actual output decomposition and its similarity to text.

\textbf{Method.}
We collect two separate features from the model: the final-layer activation map $Z(I)$ with dimensionality $C \times H^\prime \times W^\prime$, and the per-head contributions from each of the $K=H^\prime W^\prime$ image patches $\{r^h_i(I)\}_{i\in\{1, \ldots, K\}}$. We compute the per-head segmentation heatmap $L_\text{sim}(I)$ for class name $t_j$ by calculating $\langle r^h_i(I), \, M_\text{text}(t_j)\rangle$ individually for all pairs of tokens $i$ and heads $h$, and then aggregating these similarity maps such that $L_\text{sim}(I) \in \mathbb{R}^{H \times H^\prime \times W^\prime}$.

Next, we refine $Z(I)$ and $L_\text{sim}(I)$ using our neuron-attention decomposition. We select the top-$k$ neuron-head pairs by cosine similarity to class $t_j$ and denote the ordering of these pairs by $(n_1, h_1), (n_2, h_2), \ldots, (n_k, h_k)$. Then, the segmentation logits for class $t_j$ are:
\begin{equation}
    \hat{L}(I) = \sum_{r=1}^k  Z^{n_r}(I) \circ L_\text{sim}^{h_r}(I)
\end{equation}
where the superscripts $n_r$ and $h_r$ denote the top $r$-th neuron or head.

\textbf{Implementation details.}
We evaluate our method on the PASCAL Context dataset (\cite{pascal-context}). Similar to related works, we adopt a slide inference image pre-processing approach, where we specifically resize images to have a shorter side of 512 and then use a $384 \times 384$ window with a 192 stride. Additionally, we do not modify the class names in any way, both to select our top-$k$ neuron-head pairs and to compute cosine similarity.

We report all results by selecting the top $k=20000$ neuron-head pairs to text. We defer the effects of varying $k$, as well as the effect of intervening on register neurons (\cite{trained_registers}, \cite{test_time_registers}), to~\Cref{appendix-our-segmentation}.

\textbf{Main results.}
As shown in~\Cref{table-segmentation}, our method outperforms previous methods for semantic segmentation using CLIP-ResNet. To ensure fair comparison, we evaluate MaskCLIP on the same slide inference setup detailed in the previous paragraph, and use the ResNet50x16 backbone across all our experiments. We also present the results reported by SC-CLIP (\cite{sc-clip}), which is the current state-of-the-art method that uses CLIP-ViT. However, we stress that SC-CLIP leverages self-self attention, which, as shown, performs poorly when applied to CLIP-ResNet.

We present qualitative heatmaps from which we compute the segmentation maps in \Cref{table-segmentation-2}. Our neuron-attention decomposition method performs better a single aggregated segmentation map, verifying the efficacy of our method in reducing noise and focusing on the correct class.

\begin{figure}[t]
    \centering
    \begin{minipage}[c]{0.48\linewidth}
        \vspace{0mm}
        \centering
        \begin{tabular}{c c c}
            \toprule
            Method & mIoU(\%) & Backbone \\
            \midrule
            Self-self & 22.2 & RN50x16 \\ 
            MaskCLIP & 22.8 & RN50x16 \\
            \textbf{Ours} & 26.2 & RN50x16 \\
            \textcolor{gray}{SC-CLIP} & \textcolor{gray}{40.1} & \textcolor{gray}{ViT-B/16} \\
            \bottomrule
        \end{tabular}
        \captionof{table}{\textbf{Semantic segmentation performance on PASCAL Context.} For fair comparison, we implement MaskCLIP (\protect\cite{zhou-maskclip}) and $QQ^T + KK^T$ attention on the same slide inference setup we use. SC-CLIP, the current state-of-the-art, uses a ViT backbone.}
        \label{table-segmentation}
    \end{minipage}
    \hfill
    \begin{minipage}[c]{0.48\linewidth}
        \vspace{0mm}
        \centering
        \begin{tabular}{c c}
            \toprule
            Method & mIoU (\%) \\
            \midrule
            Neuron maps only & 16.5 \\
            Head maps only & 24.7 \\
            Both (multiplied) & 26.2 \\
            \bottomrule
        \end{tabular}
        \captionof{table}{\textbf{mIoU comparison of decomposition-based methods.} Segmentation by multiplying neuron features by head features performs better than by solely heads or solely neurons.}
        \label{table-segmentation-2}
    \end{minipage}
\end{figure}

\subsection{Monitoring distribution shift}
\label{Case-studies}

\begin{figure}[t]
    \centering
    \vspace{-1em}
    \centering
    \includegraphics[width=\linewidth]{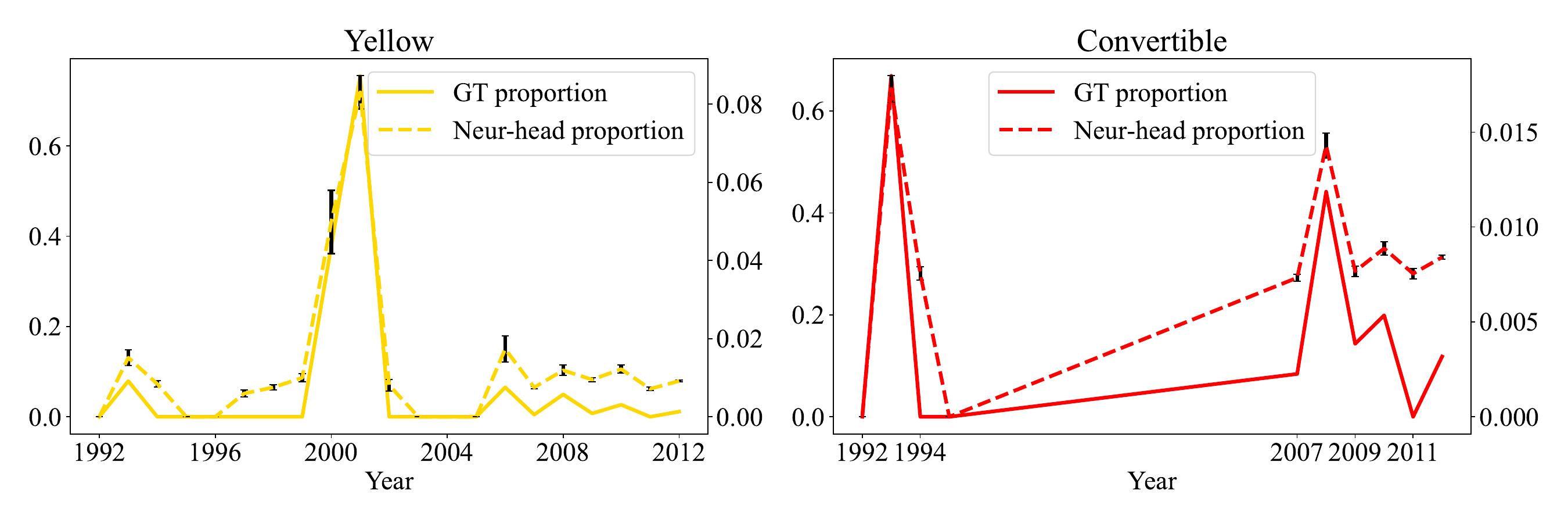}
    
    \captionof{figure}{\textbf{Monitoring the distribution shift of Stanford Cars.} We compare the ground truth concept prevalence (left y-axis) to the mean proportional contribution (in norm) of top neuron-head pairs for a given concept (right y-axis).}
    \label{fig-dist-shift}
    \vspace{-1em}
\end{figure}

As an additional application, we utilize the interpreted neurons for monitoring distribution shift between datasets. Following \cite{splice}, we consider the Stanford Cars dataset (\cite{stanford_cars}) and track the neuron-head contributions for different categories over time (see~\Cref{fig-dist-shift}). Specifically, we choose the top $k=5$ neuron-head pairs for a concept by cosine similarity, and then compute these pairs' contribution norm divided by the contribution norm of the model output itself. We compare these scores to the ground truth proportion of concepts per-year. As shown, the norms qualitatively follow similar trends to the actual distribution shift, which allows us to monitor it and summarize with text. Quantitatively, the average point-biserial correlation coefficient between the two proportions is $0.85$ for `yellow' and $0.71$ for `convertible', computed across applicable years. Notably, these concept contributions (along with tens of thousands more we do not analyze) are collected in a single forward pass per image, which makes this approach suitable for large datasets.

\section{Limitations, discussion, and future work}

We conclude by presenting two limitations of our approach and discussing future work.

\textbf{Analyzing previous layers.} 
Our approach is only applicable to the last layer of the ResNet. Earlier convolutional blocks could give us a more complete understanding of the model's computation. This is especially relevant for semantic segmentation, where related CLIP-ViT methods leverage the spatial consistency of intermediate layers for improved performance. Nevertheless, the neurons of the last layer are still useful for various downstream tasks, as shown above.

\textbf{Neuron-head pairs remain polysemantic.} 
Existing literature (\cite{vision-additive-concepts}, \cite{linear-representation}) shows evidence that models like CLIP encode concepts additively in their embedding space. While some neuron-head pairs appear less polysemantic in image retrieval examples, we qualitatively observe polysemanticity in other examples and \Cref{fig-poly}.

\textbf{Discussion and future work.}
We presented a method to analyze specific, relatively interpretable, circuits in CLIP -- neuron-head pairs. Extending and finding the correct minimal component that will be the most useful is an ongoing research question. We seek to scale and automate the assignment of fine-grained labels to such components in future work.

\textbf{Acknowledgements.} EB is grateful to XLab's Summer Research Fellowship at the University of Chicago for making this work possible. YG is supported by the Google Fellowship.

\bibliographystyle{plainnat}
\bibliography{refs}

@inproceedings{radford2021learningtransferablevisualmodels,
  title = {Learning Transferable Visual Models From Natural Language Supervision},
  author = {Radford, Alec and Kim, Jong Wook and Hallacy, Chris and Ramesh, Aditya and Goh, Gabriel and Agarwal, Sandhini and Sastry, Girish and Askell, Amanda and Mishkin, Pamela and Clark, Jack and Krueger, Gretchen and Sutskever, Ilya},
  booktitle = {Proceedings of the 38th International Conference on Machine Learning (ICML)},
  volume = {139},
  pages = {8748--8763},
  year = {2021},
  publisher = {PMLR}
}

@article{elhage2021mathematical,
   title={A Mathematical Framework for Transformer Circuits},
   author={Elhage, Nelson and Nanda, Neel and Olsson, Catherine and Henighan, Tom and Joseph, Nicholas and Mann, Ben and Askell, Amanda and Bai, Yuntao and Chen, Anna and Conerly, Tom and DasSarma, Nova and Drain, Dawn and Ganguli, Deep and Hatfield-Dodds, Zac and Hernandez, Danny and Jones, Andy and Kernion, Jackson and Lovitt, Liane and Ndousse, Kamal and Amodei, Dario and Brown, Tom and Clark, Jack and Kaplan, Jared and McCandlish, Sam and Olah, Chris},
   year={2021},
   journal={Transformer Circuits Thread},
   note={https://transformer-circuits.pub/2021/framework/index.html}
}

@inproceedings{Jia2021ScalingUV,
  title={Scaling Up Visual and Vision-Language Representation Learning With Noisy Text Supervision},
  author={Chao Jia and Yinfei Yang and Ye Xia and Yi-Ting Chen and Zarana Parekh and Hieu Pham and Quoc V. Le and Yun-Hsuan Sung and Zhen Li and Tom Duerig},
  booktitle={International Conference on Machine Learning},
  year={2021},
  url={https://api.semanticscholar.org/CorpusID:231879586}
}

@misc{zhai2023sigmoidlosslanguageimage,
      title={Sigmoid Loss for Language Image Pre-Training}, 
      author={Xiaohua Zhai and Basil Mustafa and Alexander Kolesnikov and Lucas Beyer},
      year={2023},
      eprint={2303.15343},
      archivePrefix={arXiv},
      primaryClass={cs.CV},
      url={https://arxiv.org/abs/2303.15343}, 
}

@inproceedings{he2016residual,
  author = {He, Kaiming and Zhang, Xiangyu and Ren, Shaoqing and Sun, Jian},
  booktitle = {Proceedings of 2016 IEEE Conference on Computer Vision and Pattern Recognition},
  doi = {10.1109/CVPR.2016.90},
  issn = {1063-6919},
  location = {Las Vegas, NV, USA},
  month = jun,
  pages = {770--778},
  publisher = {IEEE},
  series = {CVPR '16},
  title = {{Deep Residual Learning for Image Recognition}},
  url = {http://ieeexplore.ieee.org/document/7780459},
  year = 2016
}

@article{SecondOrderLens,
  title={Interpreting the Second-Order Effects of Neurons in CLIP},
  author={Yossi Gandelsman and Alexei A. Efros and Jacob Steinhardt},
  journal={ArXiv},
  year={2024},
  volume={abs/2406.04341},
  url={https://api.semanticscholar.org/CorpusID:270285780}
}

@inproceedings{ImageNet,
  author = {Deng, Jia and Dong, Wei and Socher, Richard and Li, Li-Jia and Li, Kai and Fei-Fei, Li},
  booktitle = {Computer Vision and Pattern Recognition, 2009. CVPR 2009. IEEE Conference on},
  organization = {IEEE},
  pages = {248--255},
  title = {Imagenet: A large-scale hierarchical image database},
  url = {https://ieeexplore.ieee.org/abstract/document/5206848/},
  year = 2009
}

@inproceedings{llava,
 author = {Liu, Haotian and Li, Chunyuan and Wu, Qingyang and Lee, Yong Jae},
 booktitle = {Advances in Neural Information Processing Systems},
 editor = {A. Oh and T. Naumann and A. Globerson and K. Saenko and M. Hardt and S. Levine},
 pages = {34892--34916},
 publisher = {Curran Associates, Inc.},
 title = {Visual Instruction Tuning},
 url = {https://proceedings.neurips.cc/paper_files/paper/2023/file/6dcf277ea32ce3288914faf369fe6de0-Paper-Conference.pdf},
 volume = {36},
 year = {2023}
}

@misc{ramesh2021zeroshottexttoimagegeneration,
      title={Zero-Shot Text-to-Image Generation}, 
      author={Aditya Ramesh and Mikhail Pavlov and Gabriel Goh and Scott Gray and Chelsea Voss and Alec Radford and Mark Chen and Ilya Sutskever},
      year={2021},
      eprint={2102.12092},
      archivePrefix={arXiv},
      primaryClass={cs.CV},
      url={https://arxiv.org/abs/2102.12092}, 
}

@inproceedings{rombach2022high,
  title={High-resolution image synthesis with latent diffusion models},
  author={Rombach, Robin and Blattmann, Andreas and Lorenz, Dominik and Esser, Patrick and Ommer, Bj{\"o}rn},
  booktitle={Proceedings of the IEEE/CVF conference on computer vision and pattern recognition},
  pages={10684--10695},
  year={2022}
}

@INPROCEEDINGS{pointclip,
  author={Zhu, Xiangyang and Zhang, Renrui and He, Bowei and Guo, Ziyu and Zeng, Ziyao and Qin, Zipeng and Zhang, Shanghang and Gao, Peng},
  booktitle={2023 IEEE/CVF International Conference on Computer Vision (ICCV)}, 
  title={PointCLIP V2: Prompting CLIP and GPT for Powerful 3D Open-world Learning}, 
  year={2023},
  volume={},
  number={},
  pages={2639-2650},
  doi={10.1109/ICCV51070.2023.00249}}

@misc{wortsman2023openclip,
  author       = {Mitchell Wortsman},
  title        = {Reaching 80\% zero-shot accuracy with OpenCLIP: ViT-g/14 trained on LAION-2B},
  year         = {2023},
  howpublished = {\url{https://laion.ai/blog/giant-openclip/}},
  note         = {Accessed: 2025-08-12}
}

@misc{textspan,
      title={Interpreting CLIP's Image Representation via Text-Based Decomposition}, 
      author={Yossi Gandelsman and Alexei A. Efros and Jacob Steinhardt},
      year={2024},
      eprint={2310.05916},
      archivePrefix={arXiv},
      primaryClass={cs.CV},
      url={https://arxiv.org/abs/2310.05916}, 
}

@inproceedings{splice,
 author = {Bhalla, Usha and Oesterling, Alex and Srinivas, Suraj and Calmon, Flavio P. and Lakkaraju, Himabindu},
 booktitle = {Advances in Neural Information Processing Systems},
 editor = {A. Globerson and L. Mackey and D. Belgrave and A. Fan and U. Paquet and J. Tomczak and C. Zhang},
 pages = {84298--84328},
 publisher = {Curran Associates, Inc.},
 title = {Interpreting CLIP with Sparse Linear Concept Embeddings (SpLiCE)},
 url = {https://proceedings.neurips.cc/paper_files/paper/2024/file/996bef37d8a638f37bdfcac2789e835d-Paper-Conference.pdf},
 volume = {37},
 year = {2024}
}

@misc{quantifying_structure,
      title={Quantifying Structure in CLIP Embeddings: A Statistical Framework for Concept Interpretation}, 
      author={Jitian Zhao and Chenghui Li and Frederic Sala and Karl Rohe},
      year={2025},
      eprint={2506.13831},
      archivePrefix={arXiv},
      primaryClass={cs.LG},
      url={https://arxiv.org/abs/2506.13831}, 
}

@inproceedings{sclip,
author = {Wang, Feng and Mei, Jieru and Yuille, Alan},
title = {SCLIP: Rethinking Self-Attention for Dense Vision-Language Inference},
year = {2024},
isbn = {978-3-031-72663-7},
publisher = {Springer-Verlag},
address = {Berlin, Heidelberg},
url = {https://doi.org/10.1007/978-3-031-72664-4_18},
doi = {10.1007/978-3-031-72664-4_18},
booktitle = {Computer Vision – ECCV 2024: 18th European Conference, Milan, Italy, September 29–October 4, 2024, Proceedings, Part XXI},
pages = {315–332},
numpages = {18},
location = {Milan, Italy}
}

@article{gemclip,
  title={Grounding Everything: Emerging Localization Properties in Vision-Language Transformers},
  author={Walid Bousselham and Felix Petersen and Vittorio Ferrari and Hilde Kuehne},
  journal={arXiv preprint arXiv:2312.00878},
  year={2023}
}

@inproceedings{cliptrase,
  title={Explore the potential of clip for training-free open vocabulary semantic segmentation},
  author={Shao, Tong and Tian, Zhuotao and Zhao, Hang and Su, Jingyong},
  booktitle={European Conference on Computer Vision},
  pages={139--156},
  year={2024},
  organization={Springer}
}

@article{sc-clip,
  title={Self-calibrated clip for training-free open-vocabulary segmentation},
  author={Bai, Sule and Liu, Yong and Han, Yifei and Zhang, Haoji and Tang, Yansong},
  journal={arXiv preprint arXiv:2411.15869},
  year={2024}
}

@article{circuits-thread:,
  author = {Cammarata, Nick and Carter, Shan and Goh, Gabriel and Olah, Chris and Petrov, Michael and Schubert, Ludwig and Voss, Chelsea and Egan, Ben and Lim, Swee Kiat},
  title = {Thread: Circuits},
  journal = {Distill},
  year = {2020},
  note = {https://distill.pub/2020/circuits},
  doi = {10.23915/distill.00024}
}

@InProceedings{network-dissection,
author = {Bau, David and Zhou, Bolei and Khosla, Aditya and Oliva, Aude and Torralba, Antonio},
title = {Network Dissection: Quantifying Interpretability of Deep Visual Representations},
booktitle = {Proceedings of the IEEE Conference on Computer Vision and Pattern Recognition (CVPR)},
month = {July},
year = {2017}
}

@misc{bills2023language,
 title={Language models can explain neurons in language models},
 author={
    Bills, Steven and Cammarata, Nick and Mossing, Dan and Tillman, Henk and Gao, Leo and Goh, Gabriel and Sutskever, Ilya and Leike, Jan and Wu, Jeff and Saunders, William
 },
 year={2023},
 howpublished = {\url{https://openaipublic.blob.core.windows.net/neuron-explainer/paper/index.html}}
}

@article{milan,
  author       = {Evan Hernandez and
                  Sarah Schwettmann and
                  David Bau and
                  Teona Bagashvili and
                  Antonio Torralba and
                  Jacob Andreas},
  title        = {Natural Language Descriptions of Deep Visual Features},
  journal      = {CoRR},
  volume       = {abs/2201.11114},
  year         = {2022},
  url          = {https://arxiv.org/abs/2201.11114},
  eprinttype    = {arXiv},
  eprint       = {2201.11114},
  bibsource    = {dblp computer science bibliography, https://dblp.org}
}

@article{LE,
  title={Linear explanations for individual neurons},
  author={Oikarinen, Tuomas and Weng, Tsui-Wei},
  journal={arXiv preprint arXiv:2405.06855},
  year={2024}
}

@inproceedings{zhou-maskclip,
  title={Extract free dense labels from clip},
  author={Zhou, Chong and Loy, Chen Change and Dai, Bo},
  booktitle={European conference on computer vision},
  pages={696--712},
  year={2022},
  organization={Springer}
}

@article{nanda2023progressgrok,
  title={Progress measures for grokking via mechanistic interpretability},
  author={Nanda, Neel and Chan, Lawrence and Lieberum, Tom and Smith, Jess and Steinhardt, Jacob},
  journal={arXiv preprint arXiv:2301.05217},
  year={2023}
}

@article{Pati1993OrthogonalMP,
  title={Orthogonal matching pursuit: recursive function approximation with applications to wavelet decomposition},
  author={Yagyensh C. Pati and Ramin Rezaiifar and Perinkulam S. Krishnaprasad},
  journal={Proceedings of 27th Asilomar Conference on Signals, Systems and Computers},
  year={1993},
  pages={40-44 vol.1},
  url={https://api.semanticscholar.org/CorpusID:16513805}
}

@InProceedings{pascal-context,
 author       = {Roozbeh Mottaghi and Xianjie Chen and Xiaobai Liu and Nam-Gyu Cho and Seong-Whan Lee and Sanja Fidler and Raquel Urtasun and Alan Yuille},
 title        = {The Role of Context for Object Detection and Semantic Segmentation in the Wild},
 booktitle    = {IEEE Conference on Computer Vision and Pattern Recognition (CVPR)},
 year         = {2014},
}

@article{LLMbio,
  author={Lindsey, Jack and Gurnee, Wes and Ameisen, Emmanuel and Chen, Brian and Pearce, Adam and Turner, Nicholas L. and Citro, Craig and Abrahams, David and Carter, Shan and Hosmer, Basil and Marcus, Jonathan and Sklar, Michael and Templeton, Adly and Bricken, Trenton and McDougall, Callum and Cunningham, Hoagy and Henighan, Thomas and Jermyn, Adam and Jones, Andy and Persic, Andrew and Qi, Zhenyi and Thompson, T. Ben and Zimmerman, Sam and Rivoire, Kelley and Conerly, Thomas and Olah, Chris and Batson, Joshua},
  title={On the Biology of a Large Language Model},
  journal={Transformer Circuits Thread},
  year={2025},
  url={https://transformer-circuits.pub/2025/attribution-graphs/biology.html}
}

@misc{incidental-poly,
      title={What Causes Polysemanticity? An Alternative Origin Story of Mixed Selectivity from Incidental Causes}, 
      author={Victor Lecomte and Kushal Thaman and Rylan Schaeffer and Naomi Bashkansky and Trevor Chow and Sanmi Koyejo},
      year={2024},
      eprint={2312.03096},
      archivePrefix={arXiv},
      primaryClass={cs.LG},
      url={https://arxiv.org/abs/2312.03096}, 
}

@misc{sharkey2025openproblemsmechanisticinterpretability,
      title={Open Problems in Mechanistic Interpretability}, 
      author={Lee Sharkey and Bilal Chughtai and Joshua Batson and Jack Lindsey and Jeff Wu and Lucius Bushnaq and Nicholas Goldowsky-Dill and Stefan Heimersheim and Alejandro Ortega and Joseph Bloom and Stella Biderman and Adria Garriga-Alonso and Arthur Conmy and Neel Nanda and Jessica Rumbelow and Martin Wattenberg and Nandi Schoots and Joseph Miller and Eric J. Michaud and Stephen Casper and Max Tegmark and William Saunders and David Bau and Eric Todd and Atticus Geiger and Mor Geva and Jesse Hoogland and Daniel Murfet and Tom McGrath},
      year={2025},
      eprint={2501.16496},
      archivePrefix={arXiv},
      primaryClass={cs.LG},
      url={https://arxiv.org/abs/2501.16496}, 
}

@article{guillaumin2014imagenet,
  title={Imagenet auto-annotation with segmentation propagation},
  author={Guillaumin, Matthieu and K{\"u}ttel, Daniel and Ferrari, Vittorio},
  journal={International Journal of Computer Vision},
  volume={110},
  number={3},
  pages={328--348},
  year={2014},
  publisher={Springer}
}

@article{helbling2025conceptattention,
  title={Conceptattention: Diffusion transformers learn highly interpretable features},
  author={Helbling, Alec and Meral, Tuna Han Salih and Hoover, Ben and Yanardag, Pinar and Chau, Duen Horng},
  journal={arXiv preprint arXiv:2502.04320},
  year={2025}
}

@InProceedings{stanford_cars,
author = {Krause, Jonathan and Stark, Michael and Deng, Jia and Fei-Fei, Li},
title = {3D Object Representations for Fine-Grained Categorization},
booktitle = {Proceedings of the IEEE International Conference on Computer Vision (ICCV) Workshops},
month = {June},
year = {2013}
}

@misc{test_time_registers,
      title={Vision Transformers Don't Need Trained Registers}, 
      author={Nick Jiang and Amil Dravid and Alexei Efros and Yossi Gandelsman},
      year={2025},
      eprint={2506.08010},
      archivePrefix={arXiv},
      primaryClass={cs.CV},
      url={https://arxiv.org/abs/2506.08010}, 
}

@article{trained_registers,
  title={Vision transformers need registers},
  author={Darcet, Timoth{\'e}e and Oquab, Maxime and Mairal, Julien and Bojanowski, Piotr},
  journal={arXiv preprint arXiv:2309.16588},
  year={2023}
}

@misc{linear-representation,
      title={The Linear Representation Hypothesis and the Geometry of Large Language Models}, 
      author={Kiho Park and Yo Joong Choe and Victor Veitch},
      year={2024},
      eprint={2311.03658},
      archivePrefix={arXiv},
      primaryClass={cs.CL},
      url={https://arxiv.org/abs/2311.03658}, 
}

@misc{vision-additive-concepts,
      title={When and why vision-language models behave like bags-of-words, and what to do about it?}, 
      author={Mert Yuksekgonul and Federico Bianchi and Pratyusha Kalluri and Dan Jurafsky and James Zou},
      year={2023},
      eprint={2210.01936},
      archivePrefix={arXiv},
      primaryClass={cs.CV},
      url={https://arxiv.org/abs/2210.01936}, 
}

@misc{xiao2024efficientstreaminglanguagemodels,
      title={Efficient Streaming Language Models with Attention Sinks}, 
      author={Guangxuan Xiao and Yuandong Tian and Beidi Chen and Song Han and Mike Lewis},
      year={2024},
      eprint={2309.17453},
      archivePrefix={arXiv},
      primaryClass={cs.CL},
      url={https://arxiv.org/abs/2309.17453}, 
}

\newpage

\appendix

\section{Appendix}

\begin{figure}[t]
    \centering
    \begin{minipage}[c]{0.48\linewidth}
        \centering
        \begin{tabular}{l c}
            \toprule
            PC(s) & Accuracy (\%) \\
            \midrule
            (Baseline) & 73.9 \\
            $\{\hat{r}^{n, h}\}$ & 73.9 \\
            $\{\hat{r}^{n}_1\}$ & 71.0 \\
            $\{\hat{r}^{n}_1, \hat{r}^{n}_2\}$ & 72.8 \\
            $\{\hat{r}^{n}_1, \hat{r}^{n}_2, \hat{r}^{n}_3\}$ & 73.3 \\
            \bottomrule
        \end{tabular}
        \captionof{table}{\textbf{Accuracy for reconstruction from principal components (RN50x64).} It remains true that neuron-head pair contributions are rank-1 in the embedding space while the representations of individual neurons are not.}
        \label{appendix-table-PCs}
    \end{minipage}
    \hfill
    \begin{minipage}[c]{0.48\linewidth}
        \vspace{0pt}
        \centering
        \includegraphics[width=0.9\linewidth]{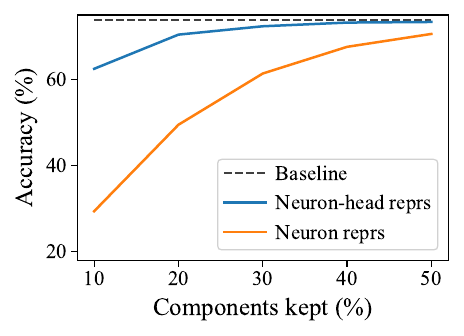}\vspace{-0.5em}
        \captionof{figure}{\textbf{Mean-ablation accuracy (RN50x64).} The same trend as \Cref{fig-mean-ablate} is shown, in which there is a sparse set of neuron-head pairs that contribute very significantly to the output value.}
        \label{appendix-fig-mean-ablate}
    \end{minipage}
\end{figure}

\subsection{Experimental results for RN50x64}
\label{appendix-other-model}
We repeat experiments from \Cref{quant-analysis} using OpenAI's CLIP-RN50x64 instead of RN50x16. As shown in \Cref{appendix-table-PCs} and \Cref{appendix-fig-mean-ablate}, neuron-head pairs and neurons display the same behavior described in the main text. Notably, the neuron representations of RN50x64 reconstruct with higher fidelity while using the same number of principal components as their RN50x16 counterparts. We also note that previous work shows that CLIP-ViT neurons are rank-1 in embedding space (\cite{SecondOrderLens}). However, for CLIP-ResNet, it is still the case that only neuron-head pairs can be approximated by a single direction in the joint embedding space.

\subsection{Reconstruction details}
\label{appendix-pca}
Specifically, we take $\hat{r}^{n, h}$ to be the first principal component computed from the top 50 $r^{n, h}(I)$ samples of $\mathcal{D}$ by norm. We compute $\{\hat{r}^n_1, \hat{r}^n_2, \ldots\}$ analogously. Additionally, we note that we add the positional embedding in this process.

In~\Cref{fig-sparse-coding} and~\Cref{table-sparse-decomposition}, rather than decomposing the $\hat{r}^n_1$ we obtain from $\mathcal{D}$, we decompose the first principle component obtained from the \textit{top 100} neuron contribution samples by norm selected  from \textit{5000} images from ImageNet test.

\subsection{Finding sub-concept neuron-head pairs.}
\label{appendix-subconcept}
We observe that our text-based decomposition doesn't capture fine-grained sub-concepts (e.g., there is no `clothing'-like description in the decomposition of the `butterfly clothing' pair $\#(624, 21)$), and that these sub-concepts are difficult to identify via cosine similarity; for instance, $\langle \hat{r}^{624,21}, \, M_\text{text}(\text{``butterfly"})\rangle >> \langle \hat{r}^{624,21}, \, M_\text{text}(\text{``butterfly clothing"})\rangle$.

\begin{figure}[t]
    \centering
    \centering
    \includegraphics[width=\linewidth]{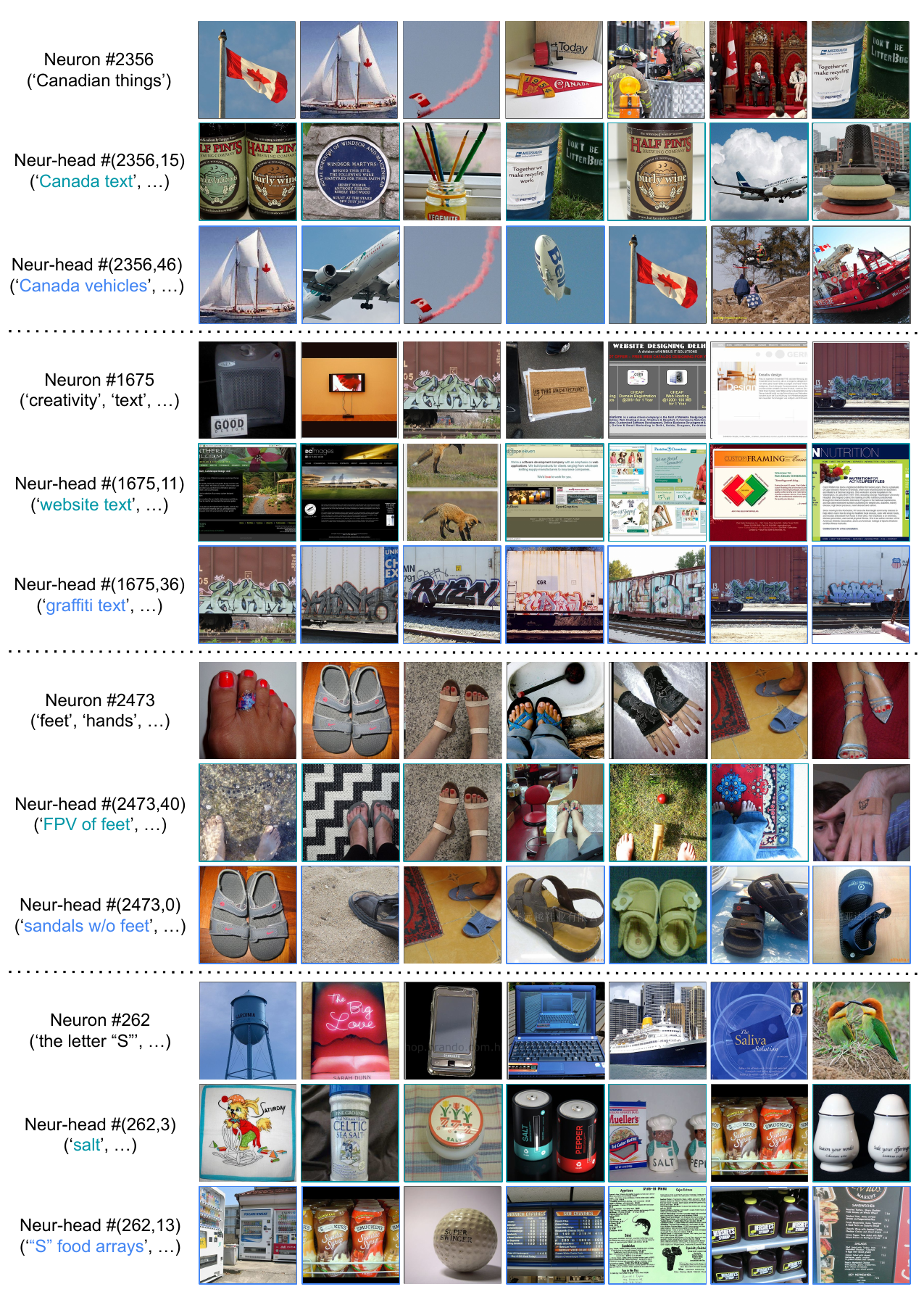}
    \captionof{figure}{\textbf{Remaining sub-concept examples.}}
    \label{fig-appendix-subconcept}
\end{figure}

Instead, we manually inspect six different neurons that are randomly selected. We narrow the search space to a pool of thirty neurons whose directions $\hat{r}^n_1$ have cosine similarity to any text embedding from the top 30k English words above a certain threshold. We automate this by prompting ChatGPT to compare top images for a pair $(n, h)$ and the top images for neuron $n$ and to output sub-concept decisions (if the sub-concept relationship exists) and descriptions of the sub-concept. We then inspect $\sim5$ of $48$ attention head pairs based on the output descriptions and present findings in~\Cref{fig-appendix-subconcept}.

\subsection{Related work: semantic segmentation using CLIP}
\label{appendix-segmentation}

Many recent works use the inherent semantic alignment of CLIP's image and text encoders for dense segmentation tasks, in which CLIP features are either processed as pseudo-labels to train another network or are used themselves for segmentation predictions (\cite{zhou-maskclip}). We focus on the latter training-free paradigm, where state-of-the-art methods incorporate a \textit{self-self attention} mechanism (\cite{gemclip}, \cite{sclip}, \cite{cliptrase}, \cite{sc-clip}). In contrast to self-self attention methods that discard the class token and rewrite attention, our proposed segmentation method uses the (decomposed) class token and does not rearrange any internal mechanisms.

In comparison to CLIP-ViT, CLIP-ResNet has disadvantages for segmentation purposes: it has more aggressive spatial downsampling (a factor of 32 compared to 16) and, as shown earlier, is not conducive to self-self attention. However, CLIP-ResNet is uniquely able to process arbitrarily-sized images without resizing, can be better at processing larger images, and its attention pooling structure makes our findings relevant to CLIP-ViT.

\subsection{Semantic segmentation using neuron-attention decomposition}
\label{appendix-our-segmentation}

\textbf{Effect of varying $k$.}
We observe that cosine similarity to text decreases drastically as $k$ increases, and the segmentation maps become noisier and more similar in range as $k$ increases. We base this observation from experiments conducted for $k$ $\in$ $\{1, 5, 10, 100, 500, 1000, 5000, 10000, 20000, 25000\}$. Intuitively, this finding should mean that smaller $k$ performs better on segmentation, but this is not the case: $k=100$ gives only $20.0$ mIoU while $k=20000$ gives $26.2$. Future work on thresholding techniques will likely bring improvement.

\textbf{Effect of register neurons.}
Modifying select $\textit{register neurons}$ at test time was recently proposed (\cite{test_time_registers}) as an alternative to trained registers (\cite{trained_registers}) that mitigate irregular attention patterns. We find a sparse set of register neurons for CLIP-ResNet and adopt the intervention detailed by \cite{test_time_registers} -- this intervention improves mIoU by 0.36\% for our method using $k=100$, but the advantage becomes negligible or even detrimental with higher $k$. We note that our method of selecting CLIP-ResNet's register neurons hinges on the empirical observation, discussed in the next section, that CLIP-ResNet's attention sink invariably appears in the last image token -- a better register neuron selection method may show further improvements.

\subsection{CLIP-ResNet attention sink}

\begin{figure}[H]
    \centering
    \includegraphics[width=\linewidth]{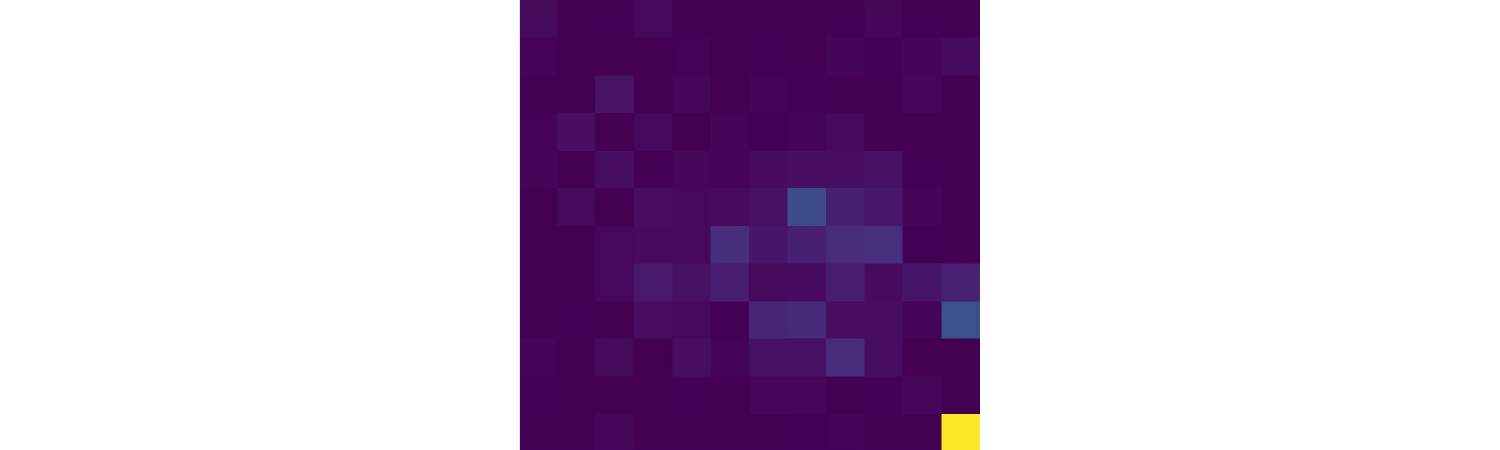}
    \captionof{figure}{\textbf{An example of CLIP-ResNet's attention sink.} We show the class token attention weights (i.e., the first row of the attention matrix), averaged over each head for an arbitary input.}
    \label{fig-appendix-sink}
    \vspace{-2em}
\end{figure}

We find that CLIP-ResNet's attention sink (\cite{xiao2024efficientstreaminglanguagemodels}) appears in the last token, such that the class token attends extremely significantly to it, as shown in~\Cref{fig-appendix-sink}.

\textbf{Finding register neurons.}
Unlike previous works on DINO and CLIP-ViT, we do not notice outlier features in the input tokens that signal an attention sink. However, we find that CLIP-ResNet's attention sink always appears in the last token. Therefore, over 1000 images from ImageNet test, we sort the neurons by the absolute value difference in the attention sink caused by intervening on each neuron (zeroing its activation). We find a sparse set of register neurons ($<18$) that causally affect the magnitude of the attention sink.


\subsection{Compute}
All experiments were run on a single A100 GPU. We note that the high amount of neuron-head pairs (for CLIP-RN50x16, $\approx 147000$) forces us to be memory-conscious (for instance, in using only 1000 images for $\mathcal{D}$).




\end{document}